# Patient-Level Diagnosis of Acute Myeloid Leukemia via Deep Learning Analysis of Bone Marrow Smear

Yuqi Ma[1†], Tianyi Wang[1†], Weihua Meng[1], Hongru Chen[2], Fajin Tao[1], Qunxian Lu[3], Lin An[3], Xiaodong Mo[2*], and Gen Yang[1*]

1 State Key Laboratory of Nuclear Physics and Technology, School of Physics, Peking University, Beijing 100871, China.

2 Peking University People's Hospital, Peking University Institute of Hematology, National Clinical Research Center for Hematologic Disease, Beijing Key Laboratory of Hematopoietic Stem Cell Transplantation, Beijing 100044, China.

3 Shanghai Dishuo Beiken Biotechnology Co., Ltd., Building 9-10, No. 3377 Kangxin Highway, Pudong New Area, Shanghai 201315, China.

†These authors have contributed equally to this work.

*Correspondence: Xiaodong Mo (mxd453@163.com) and Gen Yang (gen.yang@pku.edu.cn).

## Abstract

Bone marrow smear review remains important for acute myeloid leukemia (AML) assessment, but manual single-cell interpretation is labor-intensive and patient-level diagnosis requires aggregation of many cellular observations. We present a cell-to-patient deep learning pipeline for AML-assisted diagnosis from bone marrow smear images. The study included 258 patients from six anonymized centers, including a main cohort of 169 patients from Centers 1-3 and an external validation cohort of 89 patients from Centers 4-6. A 16-category cell annotation vocabulary was used to describe the global cellular composition, including granulocytic, monocytic, erythroid, lymphoid, eosinophilic, and other cells. Rather than identifying strict AML blasts or leukemic blasts, the model targets an expert-defined composite category termed Composite Blast-like Cells (CBLC), comprising N, N1, M, M1, R, R1, J, and J1 according to the project-wide morphological standard. A fixed YOLO-based segmentation module detected cells, predicted contours were matched to expert polygon annotations by contour IoU, and standardized single-cell crops were generated. An EfficientNet-B0 classifier was trained through a two-stage GT-to-YOLO and YOLO-to-YOLO strategy with class-imbalance correction, center-border regularization, and morphology-assisted supervision. Cell-level predictions were aggregated into patient-level CBLC ratios for AML-oriented diagnostic support. The pipeline achieved stable internal validation and maintained external generalization, with ensemble weighted F1-scores of 0.9076, 0.8696, and 0.9124 on Centers 4, 5, and 6, respectively.

## 1. Introduction

Acute myeloid leukemia (AML) diagnosis and classification require integrated evaluation of morphology, immunophenotype, cytogenetics, molecular genetics, and clinical information [1-3]. Bone marrow smear examination remains a clinically interpretable component of hematological assessment, but it relies on expert review of large numbers of cells and is affected by workload, center-specific practice, and observer variability. In addition, clinical decision support requires patient-level interpretation rather than isolated cell-level predictions [1,4].

Deep learning has enabled rapid progress in medical image analysis, including classification, detection, and segmentation tasks [5-9]. However, a practical bone marrow smear workflow must address several domain-specific challenges. Raw smear images contain variable staining, uneven background, overlapping cells, and cells at different maturation stages. Furthermore, automatically segmented cells are not identical to expert polygon annotations, creating a gap between ideal annotation-derived crops and crops generated during deployment.

We therefore developed a cell-to-patient pipeline that connects fixed cell segmentation, single-cell crop generation, binary cell classification, and patient-level aggregation. The workflow is built around two complementary descriptions of the data. First, the annotated cells are summarized using a 16-category morphological vocabulary that reflects the major cell types in the current cohort. Second, for AML-oriented diagnostic modeling, selected blast-like and early immature categories are aggregated into Composite Blast-like Cells (CBLC). The patient-level diagnostic score is then derived from the distribution of predicted CBLC-positive cells across all analyzed cells in a bone marrow smear.

## 2. Materials and Methods

### 2.1 Cohort composition and anonymization

The study used de-identified bone marrow smear images collected from six centers. As shown in Figure 1, the total cohort contained 258 patients. Centers 1-3 formed the main cohort for training and patient-level five-fold internal validation, including 169 patients, 4,023 smear images, and 46,313 annotated single cells. Centers 4-6 were reserved as the independent external validation cohort, including 89 patients, 2,045 smear images, and 22,928 annotated single cells. Across all cohorts, the dataset contained 6,068 smear images and 69,241 annotated single cells.

Healthy controls and healthy donors were uniformly described as non-AML controls. The main cohort contained 116 AML and 53 non-AML control patients, whereas the external validation cohort contained 58 AML and 31 non-AML control patients. All data used in this study were manually reviewed before modeling, and all identifiable information, including patient names, original clinical identifiers, and barcodes, was removed before analysis.

To prevent disclosure of institutional or patient identity, all centers are referred to using numeric labels only. Patient identifiers used in analysis outputs and figures were recoded with center-only prefixes, for example C1-0001, C2-0001, ..., C6-0001. No original hospital name, original patient prefix, or clinical identifier is used in the manuscript. The retrospective study was approved by the institutional ethics committee, and detailed ethics approval information will be inserted in the final public version.

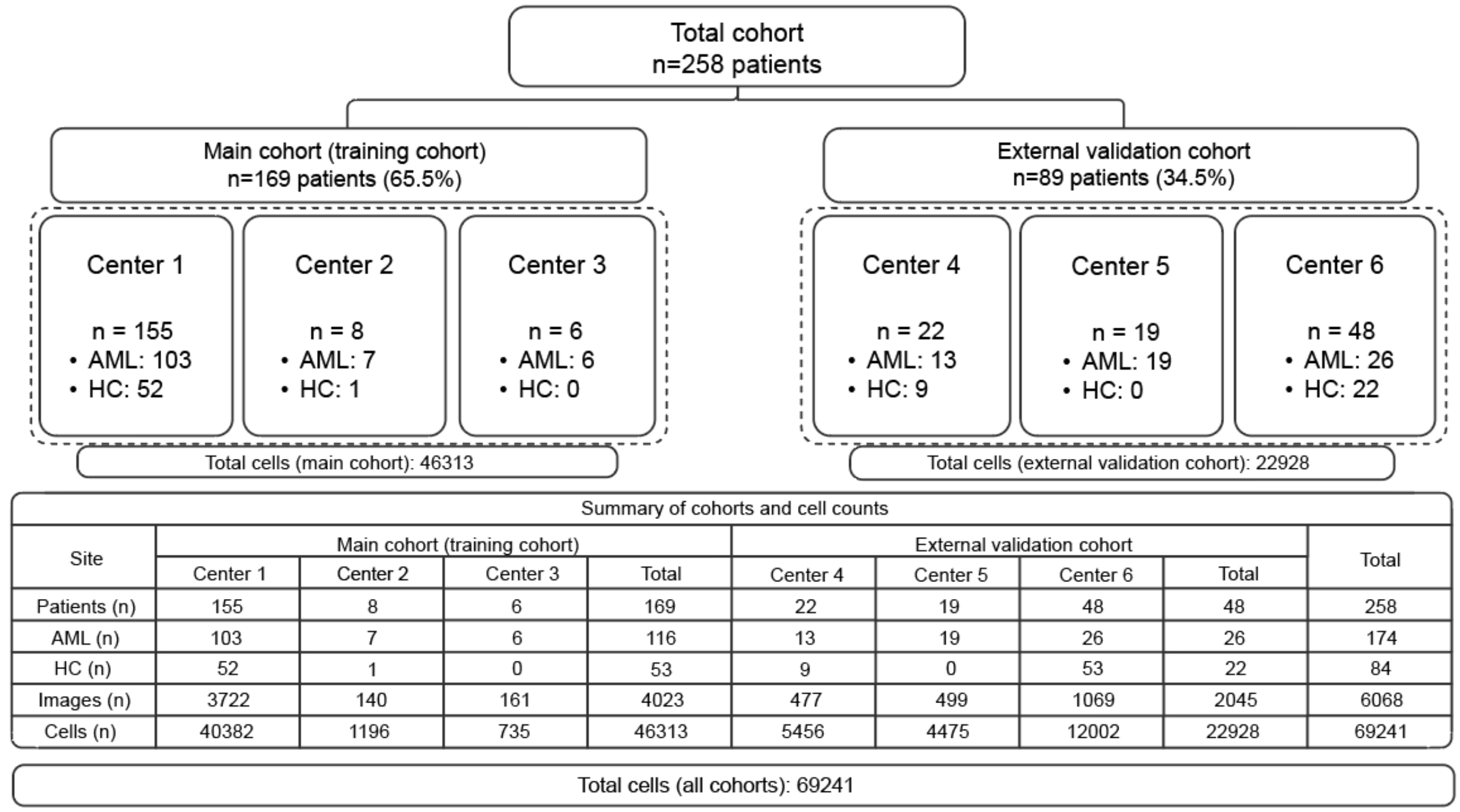

| Site | Main cohort (training cohort) | | | | External validation cohort | | | | Total |
|---|---|---|---|---|---|---|---|---|---|
| | Center 1 | Center 2 | Center 3 | Total | Center 4 | Center 5 | Center 6 | Total | |
| Patients (n) | 155 | 8 | 6 | 169 | 22 | 19 | 48 | 48 | 258 |
| AML (n) | 103 | 7 | 6 | 116 | 13 | 19 | 26 | 26 | 174 |
| HC (n) | 52 | 1 | 0 | 53 | 9 | 0 | 53 | 22 | 84 |
| Images (n) | 3722 | 140 | 161 | 4023 | 477 | 499 | 1069 | 2045 | 6068 |
| Cells (n) | 40382 | 1196 | 735 | 46313 | 5456 | 4475 | 12002 | 22928 | 69241 |

*Figure 1. Cohort composition and center-wise study split. The total cohort included 258 patients from six centers. Centers 1-3 formed the main training/internal validation cohort, and Centers 4-6 formed the external validation cohort.*

**Table 1. Cohort-level composition after data curation and center anonymization.**

| Cohort | Centers | Usage | Patients | AML / HC | Smear images | Annotated cells |
|---|---|---|---|---|---|---|
| Main cohort | Centers 1-3 | Training and internal validation | 169 | 116 / 53 | 4,023 | 46,313 |
| External validation cohort | Centers 4-6 | Independent external validation | 89 | 58 / 31 | 2,045 | 22,928 |
| Total cohort | Centers 1-6 | All study data | 258 | 174 / 84 | 6,068 | 69,241 |

## 2.2 Cell annotation vocabulary and definition of Composite Blast-like Cells

The current cohort was summarized using a 16-category cell annotation vocabulary (Figure 2): N, N1, N2, N3, N4, N5, E, M, M1, M2, R, R1, R2, R3, L, and V. These labels correspond to myeloid, monocytic, erythroid, lymphoid, eosinophilic, and other cell categories. The global label distribution was plotted as a center-wise stacked bar chart to show the contribution of the main cohort and the three external validation centers.

For patient-level AML-oriented modeling, we defined an expert-derived composite category termed Composite Blast-like Cells (CBLC). The project-wide CBLC-positive definition includes eight blast-like or early immature labels: N, N1, M, M1, R, R1, J, and J1. These labels correspond to myeloblasts, promyelocytes, monoblasts, promonocytes, proerythroblasts, early erythroblasts, megakaryoblasts, and promegakaryocytes, respectively. In the plotted 16-category distribution, the displayed CBLC-related classes are N, N1, M, M1, R, and R1; J and J1 are included in the project-wide CBLC definition when annotated but are not shown as separate bars in the 16-category global distribution. All other displayed categories were treated as non-CBLC for the binary classification task.

The patient-level leukemia probability was calculated by aggregating the predicted probabilities or binary predictions of CBLC-positive cells within each bone marrow smear and then within each patient. This label is a project-specific computational category for auxiliary diagnosis and is not equivalent to strict AML blast or leukemic blast definitions [1-4,17].

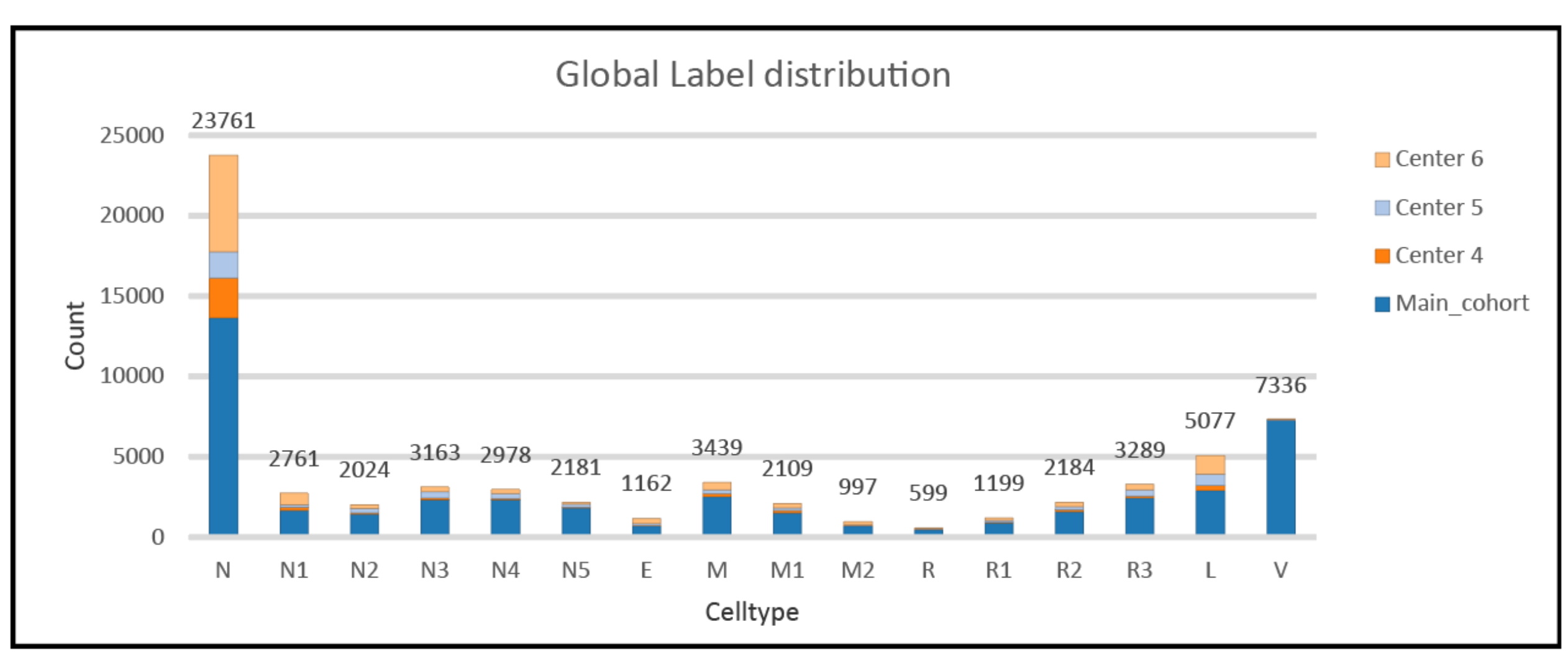

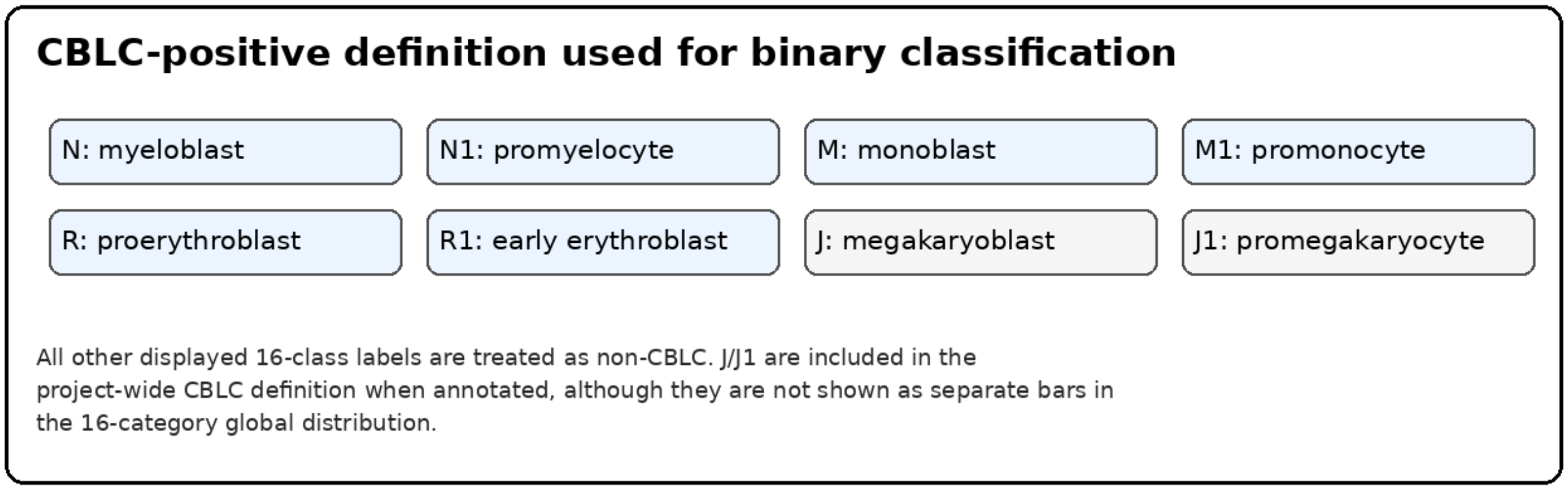

*Figure 2. Global distribution of the 16 annotated cell categories and CBLC definition. The upper panel shows center-wise stacked counts for N, N1, N2, N3, N4, N5, E, M, M1, M2, R, R1, R2, R3, L, and V. The lower panel summarizes the project-wide eight-label CBLC-positive definition used for binary classification.*

**Table 2. Sixteen-category cell annotation vocabulary and binary CBLC mapping.**

| Label | Cell category | Binary target |
|---|---|---|
| N | Myeloblast | CBLC-positive |
| N1 | Promyelocyte | CBLC-positive |
| N2 | Neutrophilic myelocyte | Non-CBLC |
| N3 | Neutrophilic metamyelocyte | Non-CBLC |
| N4 | Band neutrophil | Non-CBLC |
| N5 | Segmented neutrophil | Non-CBLC |
| E | Eosinophil | Non-CBLC |
| M | Monoblast | CBLC-positive |
| M1 | Promonocyte | CBLC-positive |
| M2 | Mature monocyte | Non-CBLC |
| R | Proerythroblast | CBLC-positive |
| R1 | Early erythroblast | CBLC-positive |
| R2 | Intermediate erythroblast | Non-CBLC |
| R3 | Late erythroblast | Non-CBLC |
| L | Lymphocyte | Non-CBLC |
| V | Other cells | Non-CBLC |

Note: In the project-wide morphological standard, J (megakaryoblast) and J1 (promegakaryocyte) are also included in the eight-label CBLC-positive definition when these labels are annotated. They are not displayed as separate bars in the 16-category distribution figure.

## 2.3 Fixed YOLO-based cell segmentation and single-cell crop generation

A fixed YOLO-based instance segmentation model was used to detect and segment cells from bone marrow smear images [10,11]. The segmentation model was not retrained within each classifier fold; instead, the same segmentation weights were applied to the main cohort and to all external validation cohorts. This design decouples segmentation evaluation from classifier cross-validation and ensures that downstream classification is evaluated under a consistent cell extraction setting.

For each detected cell, the predicted contour was matched to expert polygon annotations using contour-level intersection-over-union (IoU). Detections above the matching threshold were assigned the corresponding expert label. Cells were cropped around their center region using a high-resolution crop and resized to 224 x 224 pixels for classifier input. Edge cells and cells failing basic morphological filters, including area and circularity criteria, were excluded to reduce noisy samples. YOLO-derived single-cell crops were generated for training, validation, and external validation splits; GT-derived crops were generated for the training split only and used in the first training stage.

The overall workflow is summarized in Figure 3. Starting from de-identified bone marrow smear images, the pipeline performs cell segmentation, contour matching, single-cell crop generation, patient-level cross-validation, two-stage classifier training, cell-level inference, and patient-level aggregation of CBLC ratios for AML-assisted diagnostic interpretation.

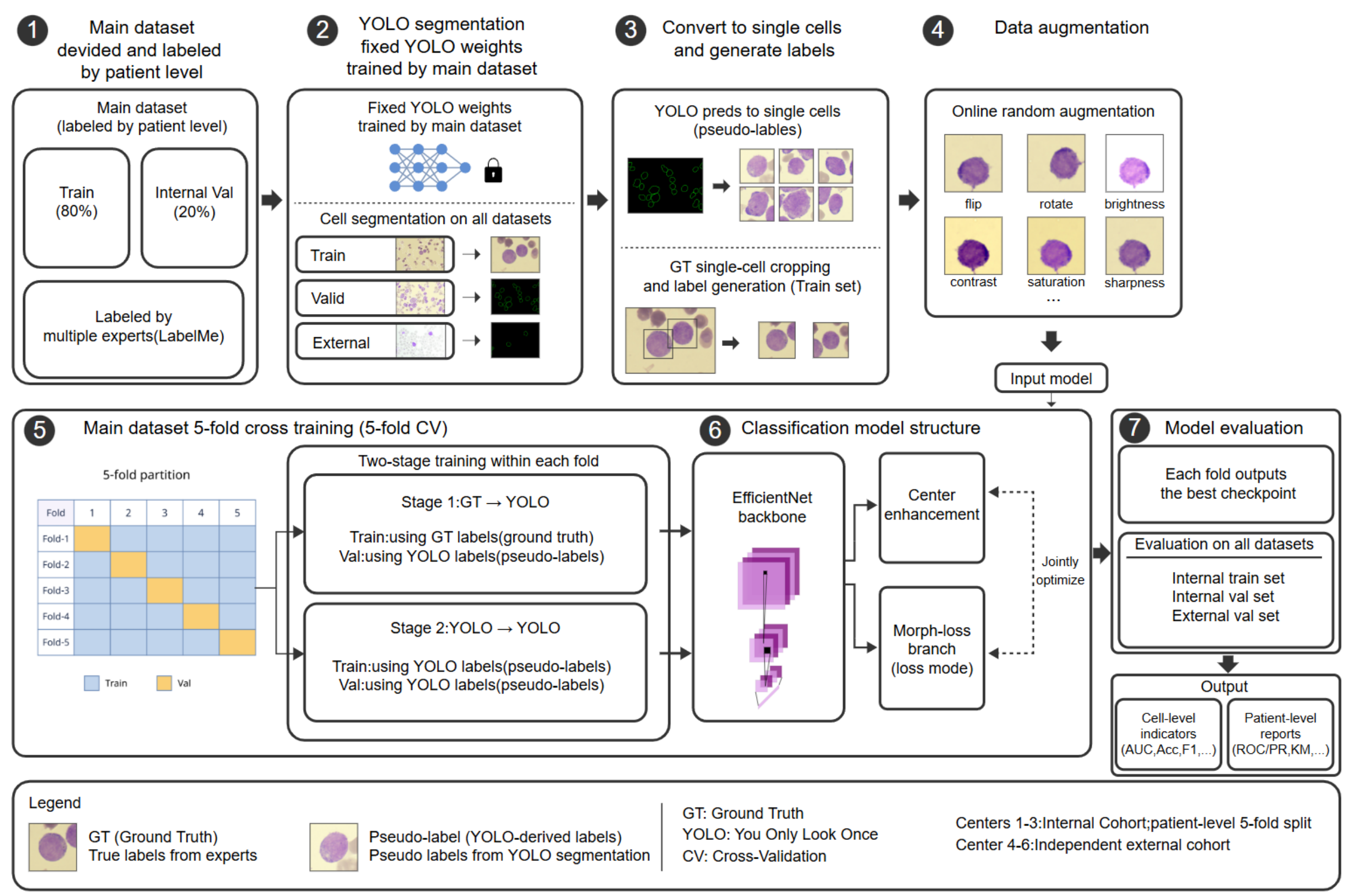

*Figure 3. Overall cell-to-patient pipeline for AML-assisted diagnosis using CBLC. The workflow links de-identified bone marrow smear images, fixed cell segmentation, IoU-based contour matching, single-cell crop generation, patient-level five-fold splitting, two-stage EfficientNet-B0 training, cell-level prediction, and patient-level aggregation.*

**Table 3. Comparison of segmentation models for bone marrow smear cell instance segmentation.**

| Segmentation model | AP50 | Precision | Recall |
|---|---|---|---|
| YOLO v11m | 0.98624 | 0.95719 | 0.96552 |
| Cellpose | 0.8574 | 0.8780 | 0.9438 |
| FastSAM | 0.7323 | 0.9338 | 0.7668 |
| InstanSeg | 0.7110 | 0.7675 | 0.8636 |
| MedSAM | 0.6279 | 0.8433 | 0.6728 |
| MobileSAM | 0.6277 | 0.8008 | 0.6928 |

The segmentation comparison included a YOLO-family model, microscopy-specialized segmentation, and SAM-derived methods [7-11]. YOLO v11m achieved the highest AP50 in the project evaluation and was therefore selected as the fixed segmentation module for the downstream pipeline.

## 2.4 Single-cell classifier and training strategy

The single-cell classifier used EfficientNet-B0 as the image encoder [12]. Input crops were normalized using ImageNet mean and standard deviation. Training augmentation included horizontal flipping, vertical flipping, rotation by 0/90/180/270 degrees, brightness and contrast perturbations, saturation perturbation, and random sharpness adjustment. The classifier head outputs two probabilities corresponding to non-CBLC and CBLC classes.

To address class imbalance, the training configuration used class-weighted loss and logit-adjusted cross-entropy with label smoothing, building on established loss and optimization strategies for imbalanced classification and deep neural network training [13,14]. The optimizer was AdamW with a learning rate of $5 \times 10^{-5}$ and weight decay of $1 \times 10^{-3}$. Training was implemented in PyTorch and PyTorch Lightning, performed for up to 50 epochs using mixed precision, and monitored by validation macro-F1 for early stopping and checkpointing [15,16].

The model included two technical refinements. First, a center-border strategy was used to encourage attention to central cell regions and reduce reliance on crop boundaries. Second, morphology-assisted supervision was incorporated through an auxiliary morphology head. The morphology targets included area, perimeter, circularity, and normalized entropy. In the current configuration, morphology was used as an auxiliary training objective rather than as direct inference-time feature fusion; therefore, enabling or disabling morphology input during testing does not change predictions when the morphology branch is used only for loss supervision.

The classifier was trained in two stages. In Stage 1, the model was trained using GT-derived single-cell crops and validated using YOLO-derived validation crops. The best Stage 1 checkpoint was then used to initialize Stage 2. In Stage 2, the model was trained using YOLO-derived training crops and validated using YOLO-derived validation crops. This GT-to-YOLO and YOLO-to-YOLO strategy allows the model to first learn clean expert-defined morphology and then adapt to segmentation-induced variations expected during inference.

## 2.5 Patient-level aggregation and evaluation metrics

After cell-level inference, predictions were grouped by anonymized patient identifier. For each patient, the workflow computed the total number of analyzed cells, number of smears, expert-derived CBLC ratio, predicted CBLC ratio, mean CBLC probability, and within-patient cell-level accuracy. The predicted CBLC ratio was used as the model-derived patient-level indicator for AML-assisted diagnostic interpretation.

Cell-level performance was evaluated using accuracy, precision, recall, macro-F1, weighted-F1, ROC-AUC, and confusion matrices. Because the binary task is class-imbalanced, weighted-F1 was used as the primary summary metric. External cohorts were never used during classifier training and were evaluated separately to assess cross-center generalization. Patient-level performance was evaluated using patient-level scatter plots and ROC curves based on aggregated CBLC ratios; cohorts without both AML and non-AML control cases were not used for ROC calculation.

## 3. Results

### 3.1 Cohort and cell-label distribution

The final study cohort included 258 patients from six anonymized centers (Figure 1). The main cohort contributed 65.5% of patients and the external validation cohort contributed 34.5% of patients. Figure 2 shows that the annotated cells were dominated by granulocytic categories, especially N, while other immature and mature categories were present at lower frequencies. This distribution supports the need for explicit class-imbalance handling during CBLC classifier training.

### 3.2 Segmentation model comparison

Among all compared segmentation approaches, YOLO v11m achieved the highest AP50 and maintained strong precision and recall (Table 3). Reliable segmentation is essential because downstream CBLC classification and patient-level aggregation depend on accurate cell localization, contour extraction, and single-cell crop quality.

### 3.3 Cell-level classification performance

The proposed pipeline achieved stable internal validation performance across patient-level five-fold validation and maintained performance on independent external validation cohorts. Because Figure 4 was removed in the revised manuscript, the fold-wise validation and external validation results are presented only in Table 4.

**Table 4. Cell-level weighted F1-score of the CBLC classification pipeline in internal validation and external validation cohorts.**

| Fold | Train | Val | Center 4 | Center 5 | Center 6 |
|---|---|---|---|---|---|
| fold1 | 0.9971 | 0.9088 | 0.9013 | 0.8477 | 0.9040 |
| fold2 | 0.9978 | 0.9101 | 0.8942 | 0.8516 | 0.9076 |
| fold3 | 0.9473 | 0.9124 | 0.9036 | 0.8746 | 0.8987 |
| fold4 | 0.9438 | 0.9341 | 0.9037 | 0.8732 | 0.8972 |
| fold5 | 1.0000 | 0.9180 | 0.9023 | 0.8633 | 0.9068 |
| ensemble | - | - | 0.9076 | 0.8696 | 0.9124 |

The ensemble weighted-F1 reached 0.9076 on Center 4, 0.8696 on Center 5, and 0.9124 on Center 6. These results suggest that the model retains useful cross-center generalization, although center-wise variation remains evident.

### 3.4 Patient-level diagnosis

Cell-level predictions were aggregated into patient-level CBLC ratios. This aggregation provides a more interpretable representation for hematological review than isolated cell-level predictions. Figure 4 combines the patient-level diagnostic scatter plot and ROC analysis. In the scatter plot, AML and non-AML control patients are displayed separately, and a model-derived CBLC-ratio threshold of 0.3468 is shown for diagnostic visualization. In the ROC analysis, the internal folds and applicable external centers achieved high patient-level AUC values. Center 5 contained AML cases without non-AML controls and therefore was not used for ROC calculation.

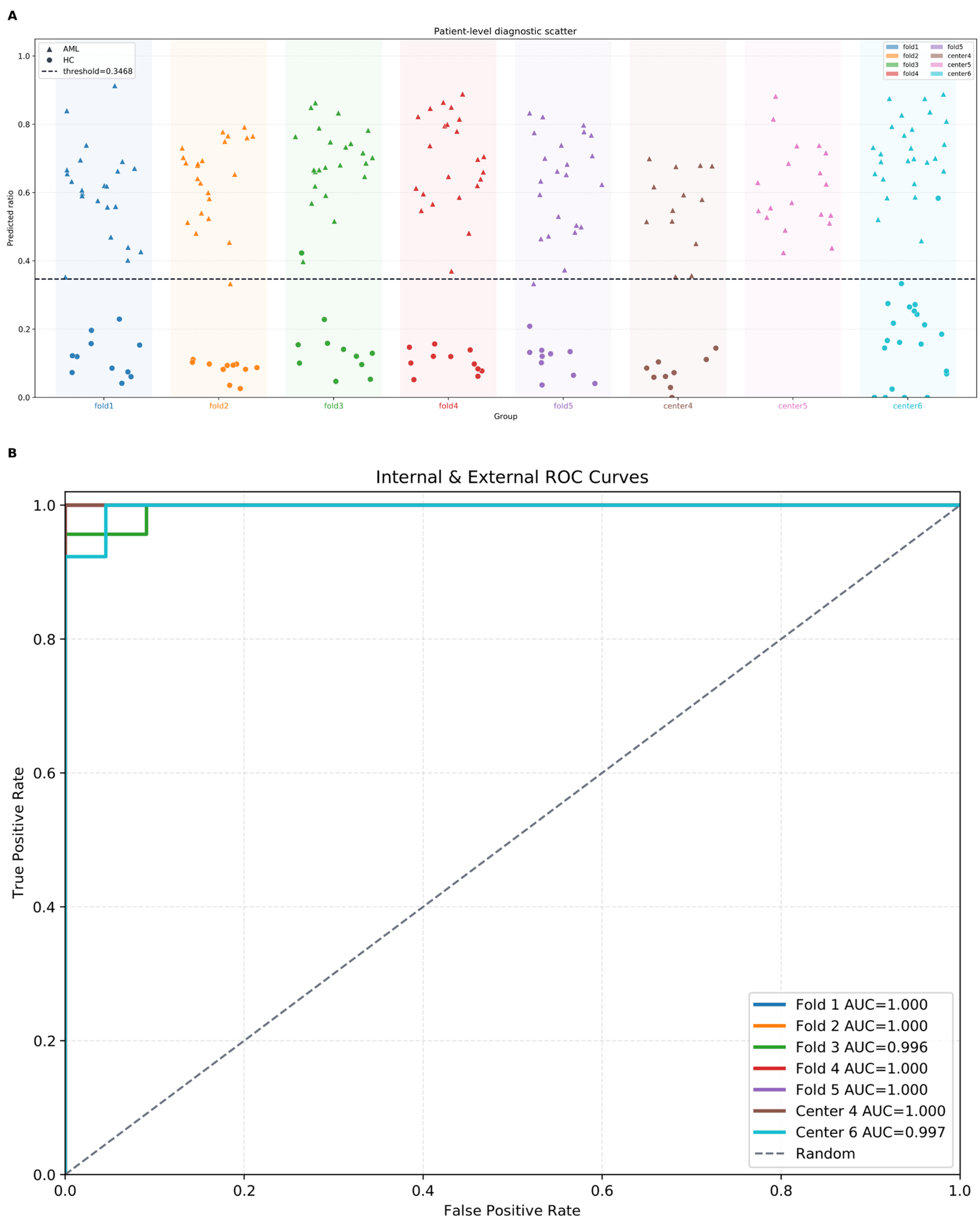

*Figure 4. Patient-level diagnostic visualization based on aggregated CBLC predictions. (A) Patient-level diagnostic scatter plot showing predicted CBLC ratios across internal folds and external validation centers, with the diagnostic visualization threshold indicated by a dashed line. (B) Internal and external ROC curves based on patient-level aggregation; Center 5 was not included in ROC analysis because it lacked non-AML control cases.*

## 4. Discussion

This study presents a practical deep learning workflow for bone marrow smear analysis that links instance-level cell segmentation to patient-level diagnostic interpretation. The pipeline is designed around the observation that clinical review depends not only on identifying individual cells but also on estimating the composition of relevant cell populations within each patient. By aggregating CBLC predictions at the patient level, the workflow produces outputs that are more interpretable for AML-assisted assessment than isolated cell-level predictions.

The revised figures clarify the relationship between the cohort structure, the cell annotation vocabulary, the CBLC definition, and patient-level diagnostic outputs. Figure 1 summarizes the six-center cohort split, Figure 2 shows the 16-category global label distribution and the eight-label CBLC-positive definition, Figure 3 illustrates the full cell-to-patient modeling workflow, and Figure 4 presents patient-level diagnostic scatter and ROC analyses. The removal of the former Figure 4 avoids redundancy because the cell-level classification performance is sufficiently summarized in Table 4.

While strict morphological definitions typically restrict blasts to the earliest developmental stages, our inclusion of early immature cells in the CBLC category is motivated by real-world hematopathological diagnostic practice [1-3,17]. For instance, in acute promyelocytic leukemia (APL, FAB M3), abnormal promyelocytes (N1) are the diagnostic hallmark [18]. Similarly, early erythroid and megakaryocytic precursors are relevant in acute erythroid leukemia and acute megakaryoblastic leukemia categories, respectively [2,3,17]. By aggregating these early precursors, our deep learning pipeline aligns more closely with diagnostic sensitivity requirements across diverse AML-related morphologies. Nevertheless, CBLC remains a computational auxiliary category and should not be interpreted as a formal clinical blast definition.

A key methodological feature is the use of fixed YOLO segmentation followed by two-stage classifier training. Direct training on expert-derived crops may produce a model that performs well on idealized cell boundaries but is less robust to automatically segmented crops. Conversely, training only on YOLO-derived crops may introduce segmentation artifacts from the beginning of training. The GT-to-YOLO and YOLO-to-YOLO strategy provides a compromise: the model first learns relatively clean expert-defined morphology and then adapts to the distribution of automatically segmented cells.

The present architecture includes center-border regularization and morphology-assisted supervision as two technical design elements. Such components are motivated by the practical observation that smear thickness, staining intensity, image acquisition, and patient composition vary substantially across centers. The morphology branch is also a promising direction because expert review relies on features such as cell size, nuclear-to-cytoplasmic ratio, chromatin pattern, cytoplasmic appearance, and granularity. In the present implementation, morphology is auxiliary rather than a direct inference input. More explicit image-morphology fusion may be explored in future work.

Several limitations should be noted. First, CBLC is a broad expert-defined computational target rather than a strict AML blast or leukemic blast category. Second, Minimal residual disease (MRD) is only a future application prospect, and this study does not include a validated MRD gold standard or clinical MRD threshold. Third, although external validation included multiple centers, larger and more diverse cohorts are needed. Finally, segmentation errors, including missed cells, overlapping cells, and edge-cell artifacts, may propagate to downstream classification and patient-level ratios.

## 5. Conclusion

We developed a cell-to-patient deep learning pipeline for AML-assisted diagnosis from bone marrow smear images using Composite Blast-like Cells. The revised manuscript aligns the text with the updated figure set: Figure 1 summarizes the six-center cohort, Figure 2 presents the 16-category cell-label distribution and CBLC definition, Figure 3 shows the full computational pipeline, Table 4 reports cell-level classification performance, and Figure 4 presents patient-level diagnostic visualization. The framework integrates fixed YOLO-based cell segmentation, IoU-guided single-cell crop generation, two-stage EfficientNet-B0 classifier training, center-border regularization, morphology-assisted supervision, and patient-level aggregation. Internal patient-level validation and external testing across anonymized centers demonstrate the feasibility of estimating patient-level CBLC ratios from single-cell predictions. The workflow provides a reproducible computational basis for AML-oriented bone marrow smear assessment, while strict blast enumeration, leukemic blast identification, and clinical MRD validation remain important directions for future work.

## References

[1] Döhner H, Wei AH, Appelbaum FR, et al. Diagnosis and management of AML in adults: 2022 recommendations from an international expert panel on behalf of the ELN. Blood. 2022;140(12):1345-1377. doi:10.1182/blood.2022016867.

[2] Khoury JD, Solary E, Abla O, et al. The 5th edition of the World Health Organization Classification of Haematolymphoid Tumours: Myeloid and Histiocytic/Dendritic Neoplasms. Leukemia. 2022;36:1703-1719. doi:10.1038/s41375-022-01613-1.

[3] Arber DA, Orazi A, Hasserjian RP, et al. International Consensus Classification of Myeloid Neoplasms and Acute Leukemias: integrating morphologic, clinical, and genomic data. Blood. 2022;140(11):1200-1228. doi:10.1182/blood.2022015850.

[4] Heuser M, Freeman SD, Ossenkoppele GJ, et al. 2021 Update on MRD in acute myeloid leukemia: a consensus document from the European LeukemiaNet MRD Working Party. Blood. 2021;138(26):2753-2767. doi:10.1182/blood.2021013626.

[5] Litjens G, Kooi T, Bejnordi BE, et al. A survey on deep learning in medical image analysis. Med Image Anal. 2017;42:60-88. doi:10.1016/j.media.2017.07.005.

[6] Esteva A, Kuprel B, Novoa RA, et al. Dermatologist-level classification of skin cancer with deep neural networks. Nature. 2017;542:115-118. doi:10.1038/nature21056.

[7] Ronneberger O, Fischer P, Brox T. U-Net: convolutional networks for biomedical image segmentation. In: Medical Image Computing and Computer-Assisted Intervention. 2015:234-241. doi:10.1007/978-3-319-24574-4_28.

[8] Stringer C, Wang T, Michaelos M, Pachitariu M. Cellpose: a generalist algorithm for cellular segmentation. Nat Methods. 2021;18:100-106. doi:10.1038/s41592-020-01018-x.

[9] Kirillov A, Mintun E, Ravi N, et al. Segment Anything. Proceedings of the IEEE/CVF International Conference on Computer Vision. 2023:4015-4026.

[10] Redmon J, Divvala S, Girshick R, Farhadi A. You Only Look Once: unified, real-time object detection. Proceedings of the IEEE Conference on Computer Vision and Pattern Recognition. 2016:779-788.

[11] Jocher G, Chaurasia A, Qiu J. Ultralytics YOLO. 2023. Available from: https://github.com/ultralytics/ultralytics.

[12] Tan M, Le QV. EfficientNet: rethinking model scaling for convolutional neural networks. Proceedings of the 36th International Conference on Machine Learning. 2019;97:6105-6114.

[13] Lin TY, Goyal P, Girshick R, He K, Dollár P. Focal loss for dense object detection. Proceedings of the IEEE International Conference on Computer Vision. 2017:2980-2988.

[14] Loshchilov I, Hutter F. Decoupled weight decay regularization. International Conference on Learning Representations. 2019.

[15] Paszke A, Gross S, Massa F, et al. PyTorch: an imperative style, high-performance deep learning library. Advances in Neural Information Processing Systems. 2019;32.

[16] Falcon WA, The PyTorch Lightning team. PyTorch Lightning. 2019. Available from: https://github.com/Lightning-AI/pytorch-lightning.

[17] Bennett JM, Catovsky D, Daniel MT, et al. Proposals for the classification of the acute leukaemias. French-American-British Cooperative Group. Br J Haematol. 1976;33(4):451-458. doi:10.1111/j.1365-2141.1976.tb03563.x.

[18] Sanz MA, Fenaux P, Tallman MS, et al. Management of acute promyelocytic leukemia: updated recommendations from an expert panel of the European LeukemiaNet. Blood. 2019;133(15):1630-1643. doi:10.1182/blood-2019-01-894980.